# Word separation in continuous sign language using isolated signs and post-processing

Razieh Rastgoo[1], Kourosh Kiani [1*], Sergio Escalera[2]

**Abstract**. Continuous Sign Language Recognition (CSLR) is a long challenging task in Computer Vision due to the difficulties in detecting the explicit boundaries between the words in a sign sentence. To deal with this challenge, we propose a two-stage model. In the first stage, the predictor model, which includes a combination of CNN, SVD, and LSTM, is trained with the isolated signs. In the second stage, we apply a post-processing algorithm to the Softmax outputs obtained from the first part of the model in order to separate the isolated signs in the continuous signs. While the proposed model is trained on the isolated sign classes with similar frame numbers, it is evaluated on the continuous sign videos with a different frame length per each isolated sign class. Due to the lack of a large dataset, including both the sign sequences and the corresponding isolated signs, two public datasets in Isolated Sign Language Recognition (ISLR), RKS-PERSIANSIGN and ASLLVD, are used for evaluation. Results of the continuous sign videos confirm the efficiency of the proposed model to deal with isolated sign boundaries detection.

1. **Introduction**

Most of the hearing-impaired people employ a sign language for communication. Information in sign language can be in the form of manual hand gestures, hand movements, body postures, and facial expressions. However, the hearing majority and also a part of the hearing-impaired community, do not know sign language. Due to the considerable amount of population of the hearing-impaired people around the world, many researchers have shown interest in developing intelligent and automatic sign language translators to facilitate the communication between the deaf community and the hearing majority. Furthermore, such translators can provide equal communication opportunities and improve public welfare for the hearing-impaired community. In line with this requirement, we propose a simple

[1] Razieh Rastgoo
Electrical and Computer Engineering Department, Semnan University, Semnan, 3513119111, Iran
E-mail: rrastgoo@semnan.ac.ir, ORCID: 0000-0001-7963-9461

[1*] Kourosh Kiani (*Corresponding Author)
Tel.: +989122361274
Fax: +98-23-33-654123
Electrical and Computer Engineering Department, Semnan University, Semnan, 3513119111, Iran
E-mail: Kourosh.kiani@semnan.ac.ir, ORCID: 0000-0001-6582-8691

[2] Sergio Escalera
Department of Mathematics and Informatics, Universitat de Barcelona, and Computer Vision Center, Barcelona, Spain
E-mail: sescalera@ub.edu, ORCID: 0000-0003-0617-8873

yet efficient model to facilitate the translation task of the continuous signs by accurately detecting the isolated signs in a continuous sign (Rastgoo, Kiani and Escalera, 2022).

Generally, Sign Language Recognition (SLR) is categorized into two groups: Isolated Sign Language Recognition (ISLR) and Continuous Sign Language Recognition (CSLR). According to the Wadhawan and Kumar (2019) study, most of the available models were proposed for ISLR (Wadhawan and Kumar, 2020). However, there are still some challenges in ISLR such as high occlusions of hands, fast hands movement, background complexity, inexistence of the large and diverse datasets, varying illumination conditions, different hand gestures, and complex interactions between hands and objects (Rastgoo, Kiani and Escalera, 2020a). In addition to the challenges in ISLR, there are also some challenges in CSLR. The most important one is detecting the isolated sign boundaries in a sign sequence. Hence, it is challenging to perform temporal segmentation on a sign sequence since there are no such explicit boundaries between the isolated signs of a sign sequence in any dataset (Nada B. Ibrahim, 2020). Another challenge is the varying length of the signing during different sentences that needs to be handled in the proposed models. In this paper, we propose a framework to solve these challenges.

Recently, Deep Learning approaches have obtained state-of-the-art performance in various tasks (Rastgoo *et al.*, 2021), especially SLR (Cui, Liu and Zhang, 2019). However, Deep Learning-based models require multiple instances of labeled sequence data to enable end-to-end sequence recognition. Since there are few public datasets for CSLR, it could be useful to employ the knowledge obtained by learning the model on isolated words to improve the training process for CSLR. In the case that the labeled sentence data is not readily available, a transfer learning mechanism can be used to facilitate the training process. Considering this, in this paper, we propose a simple yet efficient post-processing algorithm to transfer the knowledge of the trained model on the isolated signs into the CSRL problem.

The remaining paper is organized as follows. Related literature on deep models in CSLR, considering the transfer learning mechanism, is reviewed in section 2. Details of the proposed methodology are described in section 3. Results are presented and discussed in section 4. Finally, the work is discussed and concluded in section 5.

## 2. Related work

Here, we briefly review recent works in CSLR considering transfer learning. With the advent of deep learning in recent years, many recent approaches achieved state-of-the-art performance using the combination of different deep learning-based models, such as CNN and RNN (Camgoz *et al.*, 2017, 2018; Rastgoo, Kiani and Escalera, 2018, 2020b, 2020a, 2021b, 2021a, 2022; Bhagat, Vishnusai and Rathna, 2019; Gomez-Donoso, Orts-Escolano and Cazorla, 2019; Bird, Ekárt and Faria, 2020; Papastratis *et al.*, 2020; Bu *et al.*, 2020; Wadhawan and Kumar, 2020; Zhang *et al.*, 2020; Cheng *et al.*, 2020; Escobedo Cardenas and Chavez, 2020; Jiang *et al.*, 2020; Papastratis, Dimitropoulos and Daras, 2021; Rastgoo, Razieh, Kiani, Kourosh, Escalera, 2021; Rastgoo, Razieh, Kiani, Kourosh, Escalera, Sergio, Sabokrou, 2021; Sharma, Gupta and Kumar, 2021; Halvardsson *et al.*, 2021; Rastgoo, Kiani and Escalera, Sergio, Athitsos, Vassilis, Sabokrou, 2022). More specifically, while many advancements have been obtained in ISLR and CSLR with the capabilities of deep learning-based models, some challenges in both tasks still need to be discussed. For instance, the challenge of detecting the isolated sign boundaries in a continuous sign is one of them. Generally, recognizing unseen continuous signs with different sequential patterns is hard for a trained network. Furthermore, training such models is generally non-trivial, as most of them require pre-training and incorporating an iterative training strategy (Nada B. Ibrahim, 2020), which greatly lengthens the training process. Transferring the advancements obtained in ISLR into CSLR can be a useful solution for this challenge. However, some models do not consider the transfer mechanism. In this way, we briefly review recent models in two categories:

- **Transfer learning-based models**: Different transfer methodologies have been defined and used in deep learning-based models and applications, such as hand gesture recognition (Bu *et al.*, 2020), sign language recognition (Jiang *et al.*, 2020)(Bhagat, Vishnusai and Rathna, 2019; Bird, Ekárt and Faria, 2020), and speech emotion recognition (Zhang *et al.*, 2020). Using the multi-modal data for improving the recognition accuracy of a deep model is the main idea of the proposed model by Bird et al. The proposed model aimed to transfer the knowledge learned from the bigger dataset of British Sign Language (BSL) to the target model (Bird, Ekárt and Faria, 2020). In this way, two blocks are used in this model: Vision and Leap Motion. A Convolutional Neural Network and also an optimized Artificial Neural Network are employed in the first block. An evolutionary search of Artificial Neural Network topology is embedded in the second block. These blocks are fused to obtain richer features. Halvardsson et al. (Halvardsson *et al.*, 2021) transferred the knowledge of the first 20 layers of the pre-trained InceptionV3 model into a new model for static Swedish Sign Language (SSL) recognition. Furthermore, they included some new layers on top of the frozen layers, obtaining a classification accuracy of 85%. However, the results are only reported on own dataset, including 200 RGB images. In another work, Jiang et al. made various transfer learning configurations on the later layers of AlexNet using fingerspelling images from the Chinese sign language. Three training algorithms, including Adam, Stochastic Gradient Descent with Momentum (SGDM), and Root Mean Square Propagation (RMSProp), have been used in the experiments to select the most stable training algorithm. Results showed that the highest obtained recognition accuracy was 89.48% using the Adam algorithm (Jiang et al., 2020). Sharma et al. used six tri-axis accelerometers and gyroscopes, placed on both hands of the signer, to record some isolated and continuous signs. They proposed a CNN-BiLSTM-CTC network and trained on the isolated word sign dataset. They transferred the knowledge of the model to the continuous samples and analyzed the performance of the model on their own dataset. In this way, different transferring schemes, vocabulary sizes, and amount of labelled sentences have been used in the model evaluation (Sharma, Gupta and Kumar, 2021). In line with the transfer learning mechanism used in these models, in this paper, we propose a simple yet efficient post-processing model for the separation of the isolated signs into a continuous sign. We transfer the obtained knowledge from the trained model on the isolated signs into the continuous signs, relying on large datasets with a large number of samples in each class.
- **Other models**: The models in this category focus on different feature extractor models, especially deep learning-based models. Cheng et al. proposed a Fully Convolutional Network (FCN) for CSLR. To this end, a Gloss Feature Enhancement (GFE) module was proposed for sequence alignment learning. The FCN embedded in the proposed model was compared to the LSTM network and showed the superiority of the FCN in more complex real-world recognition scenarios. Results on two datasets, Chinese Sign Language (CSL) and RWTH-PHOENIX-Weather-2014 (RWTH), show that the model outperforms state-of-the-art models in the field (Cheng *et al.*, 2020). Camgoz et al. proposed a model to simultaneously consider the alignment and recognition tasks in CSLR. To this end, they defined some expert systems, namely SubUNets, to use the spatio-temporal relationships between these SubUNets for modeling the tasks. The proposed model aims to imitate human learning and educational techniques as well as feed domain-specific expert knowledge into the proposed model. Results on the RWTH-PHOENIX-Weather-2014 dataset show an accuracy improvement compared to state-of-the-art models in CSLR (Camgoz *et al.*, 2017). Papastratis et al. introduce a generative-based model for CSLR using a Generative Adversarial Network (GAN) architecture, namely Sign Language Recognition Generative Adversarial Network (SLRGAN). The spatio-temporal features extracted from video sequences are fed into the Bidirectional Long Short-Term Memory (Bi-LSTM) module of the generator to improve the recognition accuracy of the model. Furthermore, the impact of contextual information, in the form of hidden states extracted from the previous sentence, has been analyzed in the model. Results on three datasets, RWTH-Phoenix-Weather-2014, Chinese Sign Language (CSL), and Greek Sign Language (GSL) Signer Independent (SI), show relative performance improvements of 0.5 %, 0.3 %, and 1.26 %, respectively (Papastratis, Dimitropoulos and Daras, 2021).

## 3. Proposed model

In this section, we present the proposed model, which is illustrated in Figure 1.

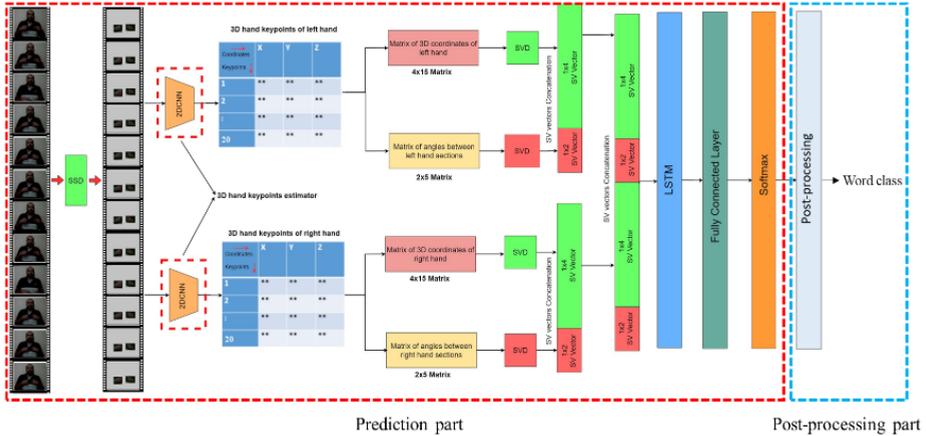

Figure 1: An overview of the proposed two-stage model. The predictor model includes a combination of CNN, SVD, and LSTM, which is trained with the isolated signs. The second stage, the post-processing part, contains a post-processing algorithm which is applied to the Softmax outputs obtained from the first part of the model to separate the isolated signs in the continuous signs.

### 3.1. Problem definition

Let $V_{train} = \{(x_{train}^1, y_{train}^1), (x_{train}^2, y_{train}^2), \cdots, (x_{train}^N, y_{train}^N)\}$ denote a set of $N$ pairs of isolated sign video $x$ and the corresponding class label $y$ of the seen data during training, with the subscript $train$ standing for training data. In a similar way, let $V_{test} = \{(x_{test}^1, y_{test}^1), (x_{test}^2, y_{test}^2), \cdots, (x_{test}^M, y_{test}^M)\}$ denote a set of $M$ pairs of isolated sign video $x$ and the corresponding class label $y$ of the unseen data during testing, with the subscript $test$ standing for test data. After the training and testing of the model, we feed a continuous sign video to the model in order to recognize and separate the isolated signs in the input sign sequence.

### 3.2. Model

Here, we describe the details of the proposed model. As stated in the previous section, the main challenge in CSLR is the boundary detection of the sign words in a continuous sign video. To solve this challenge, we propose to use transfer learning to employ the knowledge of the trained model on the isolated sign data to the continuous sign videos. More concretely, the proposed model includes the following parts:

- **The predictor model**: While the extracted features from the predictor model can be obtained using any hand-crafted or end-to-end models, we borrowed this part from our previous work in ISLR (Rastgoo, Kiani and Escalera, 2022). In this model, we use hand-crafted SVD features to feed to a LSTM Network. In order to prepare datasets for training and test the predictor model, we apply a pre-processing step to the recorded isolated sign videos with a different frame length to have equal frame numbers in all isolated sign videos. Figure 2 shows this pre-processing step.

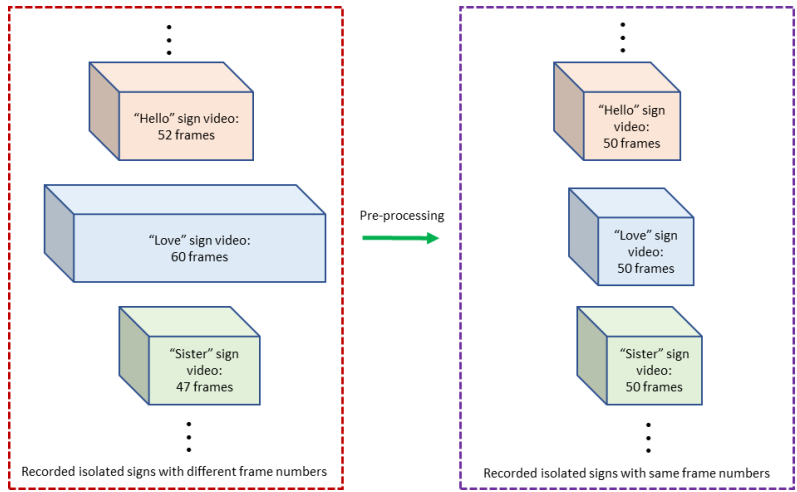

Figure 2: Equalizing the frame numbers in all isolated sign videos. This step is used in the training of the isolated sign video.

To apply the predictor model to a continuous sign video, we need the sign sequences, including the isolated signs. Due to the lack of such dataset, including both the continuous sign videos and the corresponding isolated signs, we make these continuous sign videos by concatenating isolated sign videos without any pre-processing. Figure 3 shows this step schematically.

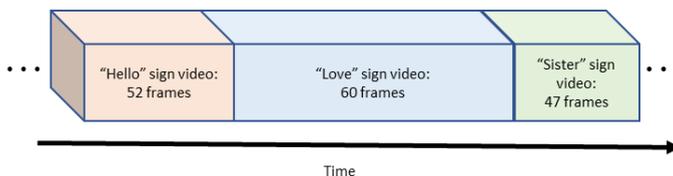

Figure 3: Building a sign sequence by concatenating the isolated signs without any pre-processing. There is a different frame length in the isolated signs.

- **Post-processing**: In this part, we apply a post-processing algorithm to the Softmax outputs obtained from the predictor model in order to separate the isolated signs in a sign sequence.

## 3.3. Details of the post-processing method

We apply a sliding window with a window size of 50 frames and feed them one by one to the hand pose estimator and also the SVD feature extractor. Then, LSTM Network (many-to-one) maps the SVD features sequence corresponding to 50 frames into a single vector. After that, this vector is passed to a Fully Connected (FC) layer. Finally, a Softmax layer is applied to the FC outputs. If the recognition accuracy of the Softmax layer for an isolated sign is higher than a predefined threshold, we accept it as a recognized isolated sign. We assign a value of 0.51 for this threshold in our experiments. The reason behind this assumption is that we generally have only one class above this threshold in each sliding window. The sliding window with stride one runs through incoming video frames.

When the first window recognizes the "Hello" sign (Figure 4), the window is slide with stride one (Figure 5). Since most of the frames in the current window are the same to the previous window, it is highly probable that the proposed model recognizes the "Hello" sign again. The proposed post-processing algorithm assigns a "Blank" label to the repeated recognized "Hello" sign. In this manner, we can separate different isolated signs.

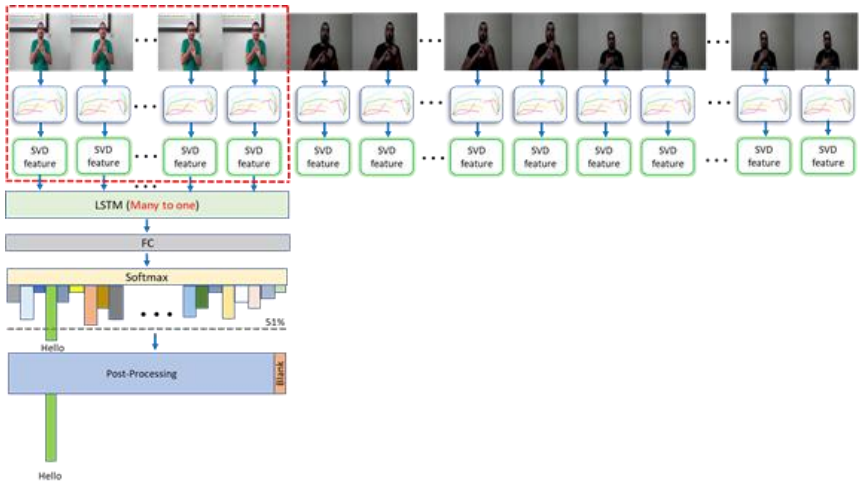

Figure 4: The first recognition of the "Hello" sign, which post-processing algorithm accepts it. The maximum value of the Softmax outputs is higher than the predefined threshold 0.51.

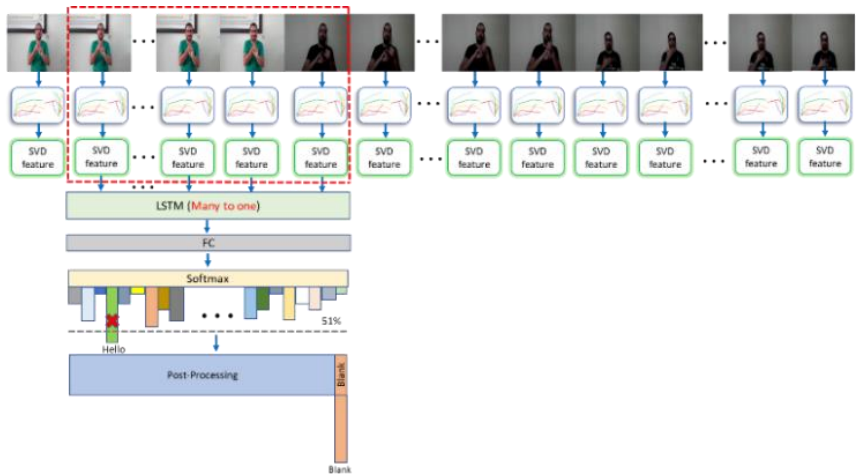

Figure 5: The immediate recognition of the "Hello" sign, which is assigned to the "Blank" label using the post-processing algorithm. Instead of having multiple similar consecutive signs in the output, the model includes multiple "Blank" labels that will be removed from the final label.

Once the recognized sign in the current window is the same as the already recognized sign class followed by only one or some "Blank" classes, we assign this recognized class to the "Blank" (Figure 6). It is worth to mentioning that in order to achieve a higher accuracy, we ignore the repeated signs.

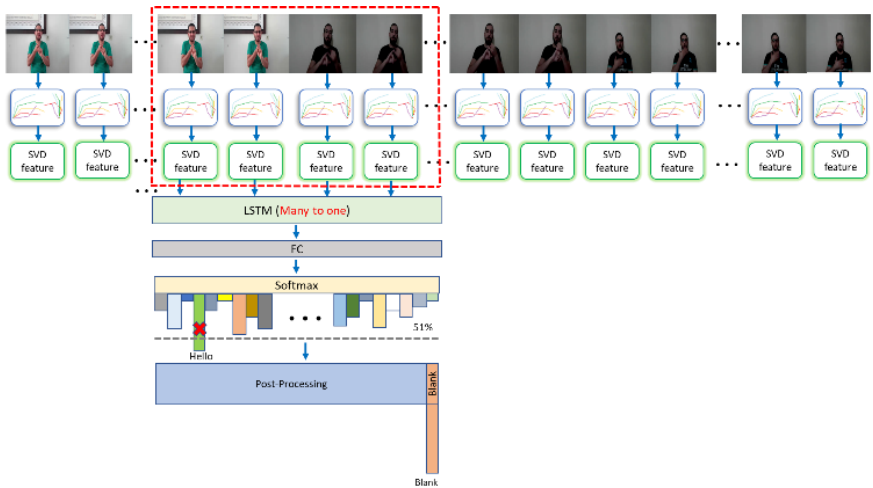

Figure 6: Assigning a "Blank" to the repeated sign class. Instead of having multiple similar consecutive signs in the output, the model includes multiple "Blank" labels that will be removed from the final label.

If all outputs of the Softmax layer are less than the predefined threshold (0.51), the post-processing algorithm assigns a "Blank" label to the current sliding window (Figure 7). This happens when the current window includes some frames from different isolated signs or unknown/untrained isolated signs.

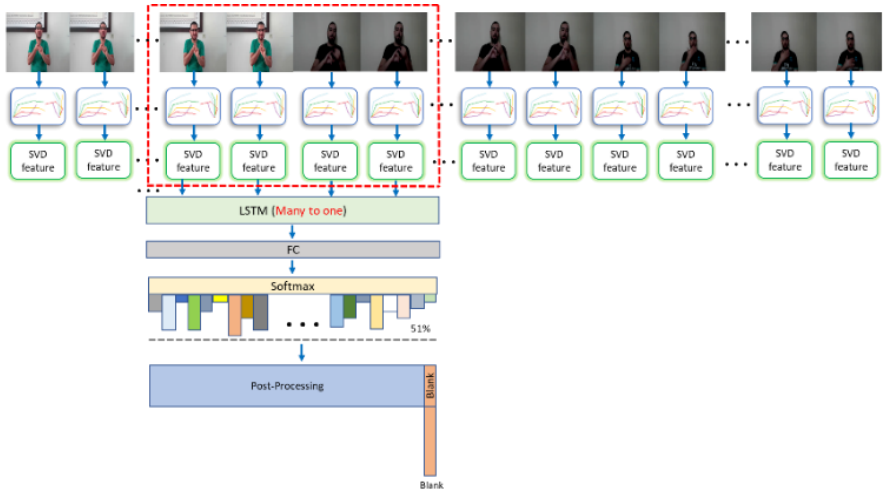

Figure 7: The post-processing algorithm assigns a "Blank" label if all outputs of the Softmax layer are less than 0.51. In this case, the proposed model cannot strongly recognize the output sign.

As sliding window is applied across the continuous video frames, the second sign class is recognized, as shown in Figure 8.

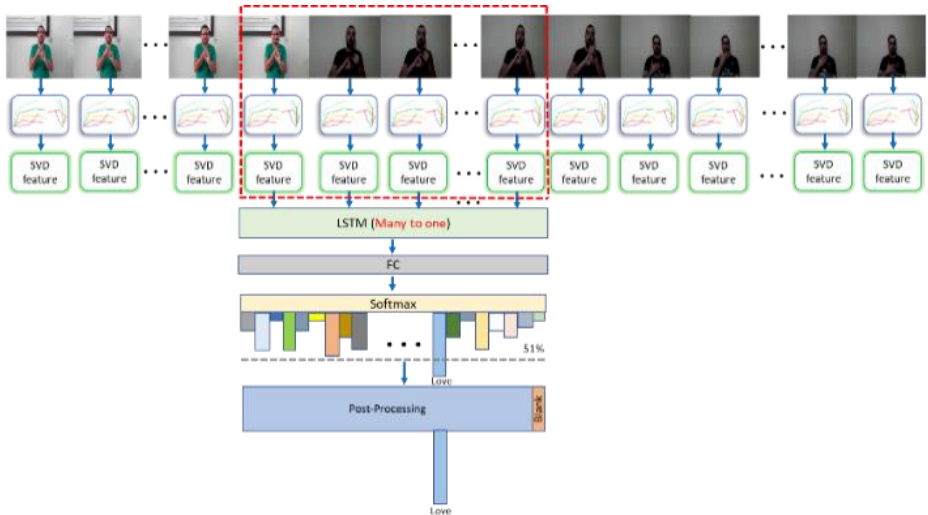

Figure 8: Recognition of the "Love" class. The maximum value of the Softmax outputs are higher than the predefined threshold 0.51.

Once an isolated sign ("Love") is recognized, the immediate recognition of the same sign ("Love") is assigned to the "Blank" class, as shown in Figure 9 (Similar to procedure of the "Hello" sign).

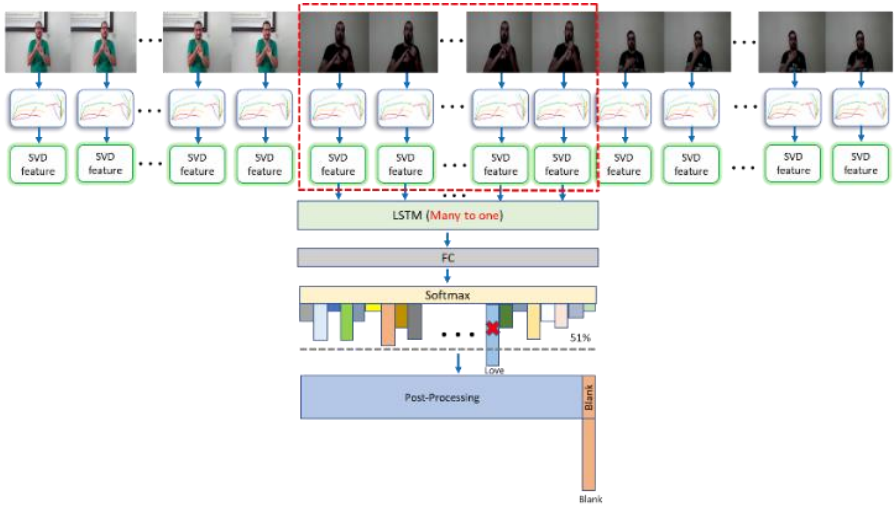

Figure 9: The immediate recognition of the "Love" sign assigned to the "Blank" label. Instead of having multiple similar consecutive signs in the output, the model includes multiple "Blank" labels that will be removed from the final label.

The sliding processing goes on and the postprocessing algorithm accepts the new sign class ("Sister"), as shown in Figure 10.

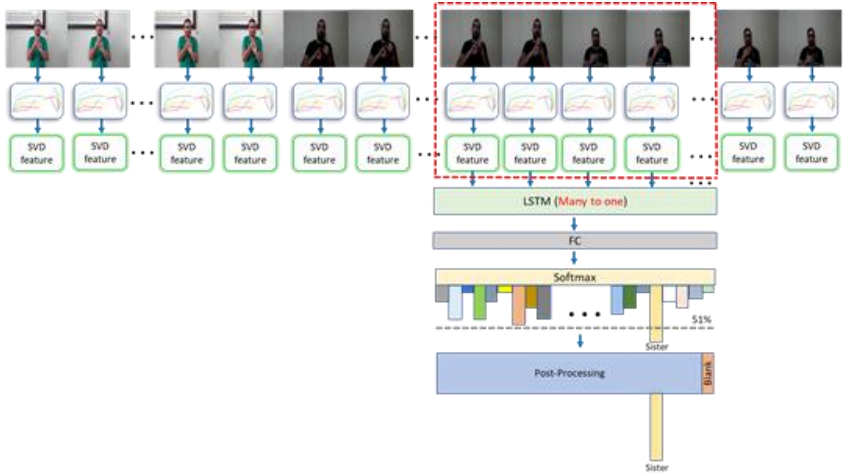

Figure 10: The new sign class ("Sister") is accepted by the post-processing algorithm.

The "Blank" is assigned to the repeated classes, as we mentioned above for other classes (Figure 11).

Figure 11: Assigning the "Blank" to the repeated sign classes. Instead of having multiple similar consecutive signs in the output, the model includes multiple "Blank" labels that will be removed from the final label.

In the final step, we only reserve one "Blank" between the recognized sign classes and remove the repeated "Blank" classes. Here, the final result is ready to convert to the text. Figure 12 shows an overview of the post-processing algorithm.

Figure 12: An overview of the post-processing algorithm. The "Blank" labels are removed from the final label.

## 4. Experimental results

In this section, we present details of the datasets and results.

### 4.1. Datasets

The available datasets in CSLR do not include both the sign streams and the corresponding isolated signs for each stream. So, to tackle this challenge, two datasets in ISLR are used for evaluation, RKS-PERSIANSIGN (Rastgoo, Kiani and Escalera, 2020a) and ASLLVD (Neidle, Thangali and Sclaroff, 2012). The first dataset includes 10'000 RGB videos of 100 Persian signs using 10 contributors, 5 women and 5 men, in a simple background with a maximum distance of 1.5 meter between the contributor and camera. There are 100 video samples for each Persian sign word. The second dataset, ASLLVD, contains some American sign words with their corresponding annotations. There are different sample numbers in each class label. We selected the 100 sign categories which contain at least seven video samples. We pre-process these datasets to have equally frame numbers for all video samples during the training phase. However, we do not perform any pre-processing on the test data and use them with a different frame length. To build continuous video signs, the un-preprocessed isolated signs with a different frame length corresponding to all classes are concatenated to each other. Each continuous video sign contains only one sample from each isolated sign class. In this way, we have 100 and 7 continuous video signs in the PERSIANSIGN and ASLLVD datasets, respectively. Table 1 shows details of datasets used in model evaluation.

Table 1: Details of datasets used in model evaluation.

|  | RKS-PERSIANSIGN | ASLLVD |
|---|---|---|
| Isolated sign video numbers per class | 100 | 7 |
| Total isolated sign video numbers | 10'000 | 700 |
| Total continuous sign videos | 100 | 7 |
| Subjects per class | 10 | - |
| Video sign language | Persian | American |

### 4.2. Implementation details

Our evaluation has been done on an Intel(R) Xeon(R) CPU E5-2699 (2 processors) with 90GB RAM with Microsoft Windows 10 operating system and Python software with NVIDIA Tesla K80 GPU. We implemented our model in the Keras library. We use *Adam* optimizer with a mini-batch size of 50. The learning rate starts from 0.005 and is divided by 10 every 10 epochs. The proposed model is trained for a total of 200 epochs with early stopping. In addition, we use a weight decay of *1e-4* and a momentum of 0.92. We used two datasets for the evaluation: RKS-PERSIANSIGN and ASLLVD, which are divided randomly into training (80%) and testing (20%) sets. Table 2 shows the details of the experimentally set model parameters.

Table 2: Details of the parameters used in the proposed model.

| Parameter | Value | Parameter | Value |
|---|---|---|---|
| Weight decay | 1e-4 | Epoch numbers | 200 |
| Learning rate | 0.005 | Number of frames per video sample | 50 |
| Batch size | 50 | Processing way | GPU |
| Keypoint dimension | 21x3 | Number of singular values | 12 |

### 4.3. Ablation analysis

In the first experiment, we report the recognition accuracy of the predictor model on the isolated signs for both datasets (Table 3). As this table shows, the predictor model accurately recognizes the isolated signs in both datasets.

Table 3: Results of the proposed model in three phases: Train, Validation, and Test.

| Phase \ Dataset | RKS-PERSIANSIGN | ASLLVD |
|---|---|---|
| Proposed model (Train) | 99.8 ± 0.04 | 95 ± 0.05 |
| Proposed model (Validation) | 99.6 ± 0.04 | 94 ± 0.05 |
| Proposed model (Test) | 99.5 ± 0.04 | 93 ± 0.05 |

To analyze the impact of the post-processing methodology, we present our ablation analysis on two datasets. As we stated in subsection 4.1, we built 100 and 7 continuous sign videos using the isolated signs in RKSPERSIANSIGN and ASLLVD datasets, respectively. Table 4 shows details of the results for one continuous sign video of the ASLLVD dataset. Since each continuous sign video includes 100 isolated signs with a different frame length, the 100 Softmax output values for each isolated sign are shown in each row of this table. Furthermore, the last column shows the maximum value of the Softmax output values for each isolated sign. Also, the last row presents the average of the Softmax output values for 100 isolated signs in a continuous sign video. As this table shows, there are some false recognition in some isolated signs. For example, in row 45 of this table, the maximum Softmax output value is assigned to the 63th sign class with the value of 0.39. However, the true class is the sign class 45 with the Softmax output value of 0.37. A similar situation happens for sign class 51, which the proposed model falsely recognizes the sign class 64. The point is that both Softmax output values for the true sign class as well as the recognized sign class are lower than the threshold 0.51. These false recognitions in the ASLLVD dataset comes from the similarities between the signs in the dataset. In the post-processing algorithm, the "Blank" label will be assigned to these isolated signs with the recognized sign class lower than the threshold 0.51.

Table 4: The first concatenated sign video of the ASLLVD dataset. There are two false recognition in this video.

| Softmax class index | 1 | 2 | … | 45 | … | 51 | … | 63 | 64 | … | 100 | Max |
|---|---|---|---|---|---|---|---|---|---|---|---|---|
| Softmax output of the sign class 1 | 0.54 | * | … | * | … | * | … | * | * | … | * | 0.54 |
| Softmax output of the sign class 2 | * | 0.56 | … | * | … | * | … | * | * | … | * | 0.56 |
| ⋮ | | | | | | ⋮ | | | | | | ⋮ |
| Softmax output of the sign class 45 | * | * | … | 0.37 | … | * | … | .39 | * | … | * | 0.39 |
| ⋮ | | | | | | ⋮ | | | | | | ⋮ |
| Softmax output of the sign class 51 | * | * | … | * | … | 0.33 | … | * | 0.35 | … | * | 0.35 |
| ⋮ | | | | | | ⋮ | | | | | | ⋮ |
| Softmax output of the sign class 100 | * | * | … | * | … | * | … | * | * | … | 0.59 | 0.59 |
| | | | | | | | | | | Average Softmax outputs of the 100 sign classes | | 0.59 |

The Softmax output values of the false recognized isolated signs as well the true sign class of the seven continuous sign videos created using ASLLVD dataset can be found in Table 5. The average of recognized Softmax outputs on ASLLVD is 0.59. It is worth mentioning that the average of the Softmax outputs is different from the recognition accuracy of the proposed model. The average of the Softmax outputs in each continuous sign video is the average of the maximum value of the Softmax outputs in each recognized isolated sign. The recognition accuracy of the proposed model for isolated sign videos and also the number of false recognized isolated signs in continuous sign videos can be found in Table 3 and 5. As Table 5 shows, there are 12 false recognized isolated signs in 700 isolated signs corresponding to 7 continuous sign videos.

Table 5: Details of the recognition accuracy of the proposed post-processing algorithm on the ASLLVD dataset.

| Concatenated sign video number | Avg of Softmax output of each sign video | Index of correct sign class | Softmax output of correct sign class | Index of false sign class | Softmax output of false sign class | Concatenated sign video number | Avg of Softmax output of each sign video | Index of correct sign class | Softmax output of correct sign class | Index of false sign class | Softmax output of false sign class |
|---|---|---|---|---|---|---|---|---|---|---|---|
| 1 | 0.59 | 45 | 0.37 | 63 | 0.39 | 5 | 0.59 | 45 | 0.37 | 63 | 0.39 |
|   |      | 51 | 0.33 | 64 | 0.35 |   |      | 64 | 0.33 | 51 | 0.35 |
| 2 | 0.59 | 8  | 0.36 | 18 | 0.38 | 6 | 0.59 | 18 | 0.34 | 8  | 0.36 |
| 3 | 0.59 | 45 | 0.38 | 63 | 0.40 |   |      | 63 | 0.35 | 45 | 0.37 |
|   |      | 50 | 0.44 | 80 | 0.45 | 7 | 0.59 | 45 | 0.33 | 63 | 0.35 |
| 4 | 0.59 | 45 | 0.37 | 63 | 0.39 |   |      | 50 | 0.46 | 80 | 0.48 |

Similar to the ASLLVD dataset, we present the details of the results for one continuous sign video of the RKSPERSIANSIGN dataset in Table 6. Each row of this table is corresponding to one isolated sign in the continuous sign video. The last column of the Table 6 shows the maximum value of the Softmax output values for each isolated sign. Also, the last row presents the average of the Softmax output values for 100 isolated signs in a continuous sign video. As one can see in Table 6, there are some false recognition in some isolated signs. For example, in row 17 of this table, the maximum Softmax output value is assigned to the 19th sign class with the value of 0.47. Although, the true class is the sign class 17 with the Softmax output value of 0.45. A similar situation happens for sign class 86, which the proposed model falsely recognizes the sign class 66. The point is that both Softmax output values for the true sign class as well as the recognized sign class are lower than the threshold 0.51. These false recognitions in the RKSPERSIANSIGN dataset comes from the similarities between the signs in the dataset. In the post-processing algorithm, the "Blank" label will be assigned to these isolated signs with the recognized sign class lower than the threshold 0.51.

Table 6: The first concatenated sign video of the RKS-PERSIANSIGN dataset.

| Softmax class index | 1 | … | 17 | … | 19 | … | 66 | … | 86 | … | 100 | Max |
|---|---|---|---|---|---|---|---|---|---|---|---|---|
| Softmax output of the sign class 1 | 0.98 | … | * | … | * | … | * | … | * | … | * | 0.98 |
| ⋮ |   | … | ⋮ | … | ⋮ | … | ⋮ | … | ⋮ | … | ⋮ | ⋮ |
| Softmax output of the sign class 17 | * | … | 0.45 | … | 0.47 | … | * | … | * | … | * | 0.47 |
| ⋮ |   | … | ⋮ | … | ⋮ | … | ⋮ | … | ⋮ | … | ⋮ | ⋮ |
| Softmax output of the sign class 86 | * | … | * | … | * | … | 0.45 | … | 0.43 | … | * | 0.45 |
| ⋮ |   | … | ⋮ | … | ⋮ | … | ⋮ | … | ⋮ | … | ⋮ | ⋮ |
| Softmax output of the sign class 100 | * | … | * | … | * | … | * | … | * | … | 0.99 | 0.99 |
| **Average Softmax outputs of the 100 sign classes** | | | | | | | | | | | | **0.97** |

In addition to this, Table 7 shows the Softmax output values of the false recognized isolated signs as well the true isolated signs of 100 continuous sign videos created using RKSPERSIANSIGN dataset. The average of recognized Softmax outputs of the isolated signs in the created continuous sign videos of the RKSPERSIANSIGN have been shown in this table. It is worth mentioning that the average of the Softmax outputs is different from the recognition accuracy of the proposed model. The recognition accuracy of the proposed model for isolated sign videos and also the number of false recognized isolated signs in continuous sign videos can be found in Table 3 and 7. In overall, there are 24 false recognized isolated signs in 10000 isolated signs corresponding to 100 continuous sign videos.

*Table 7: Details of the recognition accuracy of the proposed postprocessing algorithm on the RKS-PERSIANSIGN dataset.*

| Concatenated sign video number | Avg of Softmax output of each sign video | Index of correct sign class | Softmax output of correct sign class | Index of false sign class | Softmax output of false sign class | Concatenated sign video number | Avg of Softmax output of each sign video | Index of correct sign class | Softmax output of correct sign class | Index of false sign class | Softmax output of false sign class | Concatenated sign video number | Avg of Softmax output of each sign video | Index of correct sign class | Softmax output of correct sign class | Index of false sign class | Softmax output of false sign class |
|---|---|---|---|---|---|---|---|---|---|---|---|---|---|---|---|---|---|
| 1 | 0.97 | 17 | 0.45 | 19 | 0.47 | 34 | 0.97 | 86 | 0.44 | 66 | 0.49 | 68 | 0.98 | 63 | 0.48 | 45 | 0.49 |
|  |  | 86 | 0.43 | 66 | 0.45 |  |  | 63 | 0.46 | 45 | 0.48 | 69 | 0.99 | - | - | - | - |
| 2 | 0.98 | 19 | 0.47 | 17 | 0.49 | 35 | 0.99 | - | - | - | - | 70 | 0.99 | - | - | - | - |
| 3 | 0.98 | 63 | 0.45 | 45 | 0.49 | 36 | 0.99 | - | - | - | - | 71 | 0.99 | - | - | - | - |
| 4 | 0.99 | - | - | - | - | 37 | 0.99 | - | - | - | - | 72 | 0.99 | - | - | - | - |
| 5 | 0.99 | - | - | - | - | 38 | 0.99 | - | - | - | - | 73 | 0.99 | - | - | - | - |
| 6 | 0.99 | - | - | - | - | 39 | 0.99 | - | - | - | - | 74 | 0.99 | - | - | - | - |
| 7 | 0.98 | 45 | 0.43 | 63 | 0.45 | 40 | 0.98 | 86 | 0.48 | 66 | 0.49 | 75 | 0.99 | - | - | - | - |
| 8 | 0.99 | - | - | - | - | 41 | 0.99 | - | - | - | - | 76 | 0.99 | - | - | - | - |
| 9 | 0.99 | - | - | - | - | 42 | 0.99 | - | - | - | - | 77 | 0.97 | 63 | 0.47 | 45 | 0.48 |
| 10 | 0.97 | 19 | 0.45 | 17 | 0.49 | 43 | 0.99 | - | - | - | - |  |  | 86 | 0.48 | 66 | 0.49 |
|  |  | 45 | 0.44 | 63 | 0.49 | 44 | 0.98 | 45 | 0.45 | 63 | 0.46 | 78 | 0.99 | - | - | - | - |
| 11 | 0.99 | - | - | - | - | 45 | 0.99 | - | - | - | - | 79 | 0.99 | - | - | - | - |
| 12 | 0.99 | - | - | - | - | 46 | 0.99 | - | - | - | - | 80 | 0.99 | - | - | - | - |
| 13 | 0.99 | - | - | - | - | 47 | 0.99 | - | - | - | - | 81 | 0.99 | - | - | - | - |
| 14 | 0.99 | - | - | - | - | 48 | 0.99 | - | - | - | - | 82 | 0.99 | - | - | - | - |
| 15 | 0.98 | 63 | 0.43 | 45 | 0.49 | 49 | 0.99 | - | - | - | - | 83 | 0.99 | - | - | - | - |
| 16 | 0.99 | - | - | - | - | 50 | 0.97 | 17 | 0.48 | 19 | 0.49 | 84 | 0.99 | - | - | - | - |
| 17 | 0.99 | - | - | - | - |  |  | 63 | 0.46 | 45 | 0.48 | 85 | 0.99 | - | - | - | - |
| 18 | 0.99 | - | - | - | - | 51 | 0.99 | - | - | - | - | 86 | 0.99 | - | - | - | - |
| 19 | 0.99 | - | - | - | - | 52 | 0.99 | - | - | - | - | 87 | 0.99 | - | - | - | - |
| 20 | 0.99 | - | - | - | - | 53 | 0.99 | - | - | - | - | 88 | 0.99 | - | - | - | - |
| 21 | 0.99 | - | - | - | - | 54 | 0.99 | - | - | - | - | 89 | 0.99 | - | - | - | - |
| 22 | 0.99 | - | - | - | - | 55 | 0.99 | - | - | - | - | 90 | 0.99 | - | - | - | - |
| 23 | 0.99 | - | - | - | - | 56 | 0.99 | - | - | - | - | 91 | 0.99 | - | - | - | - |
| 24 | 0.99 | - | - | - | - | 57 | 0.99 | - | - | - | - | 92 | 0.99 | - | - | - | - |
| 25 | 0.97 | 19 | 0.46 | 17 | 0.49 | 58 | 0.99 | - | - | - | - | 93 | 0.97 | 45 | 0.45 | 63 | 0.46 |
|  |  | 45 | 0.45 | 63 | 0.47 | 59 | 0.98 | 17 | 0.46 | 19 | 0.48 |  |  | 63 | 0.45 | 45 | 0.46 |
| 26 | 0.99 | - | - | - | - | 60 | 0.99 | - | - | - | - | 94 | 0.99 | - | - | - | - |
| 27 | 0.99 | - | - | - | - | 61 | 0.99 | - | - | - | - | 95 | 0.99 | - | - | - | - |
| 28 | 0.99 | - | - | - | - | 62 | 0.99 | - | - | - | - | 96 | 0.99 | - | - | - | - |
| 29 | 0.99 | - | - | - | - | 63 | 0.99 | - | - | - | - | 97 | 0.99 | - | - | - | - |
| 30 | 0.99 | - | - | - | - | 64 | 0.99 | - | - | - | - | 98 | 0.97 | 19 | 0.48 | 17 | 0.49 |
| 31 | 0.99 | - | - | - | - | 65 | 0.99 | - | - | - | - |  |  | 63 | 0.47 | 45 | 0.48 |
| 32 | 0.99 | - | - | - | - | 66 | 0.99 | - | - | - | - | 99 | 0.99 | - | - | - | - |
| 33 | 0.99 | - | - | - | - | 67 | 0.99 | - | - | - | - | 100 | 0.99 | - | - | - | - |

## 5. Discussion and conclusion

In this work, we proposed a simple yet efficient post-processing methodology for the separation of the isolated signs in a continuous sign video, as a long challenging task in Computer Vision. Due to the lack of a continuous sign dataset including both the continuous signs and the corresponding isolated signs, we use the datasets in isolated sign language and concatenate them to make the continuous sign videos. To build continuous video signs, the un-preprocessed isolated signs with a different frame length corresponding to all classes are concatenated to each other. Each continuous video sign contains only one video from each isolated sign class. In this way, we have 100 and 7 continuous video signs in the PERSIANSIGN and ASLLVD datasets, respectively. While the proposed model is trained on the isolated signs with similar frame numbers, it is evaluated on the continuous sign videos with a different frame length per each isolated sign. Furthermore, since there is no similar work to ours, we cannot compare the proposed model with other models. The proposed methodology used a predefined threshold, 0.51, to accept or reject a recognized class in the current sliding window. The intuition behind this predefined value is that we generally can have only one class above this threshold in the current sliding window. This comes from the property of the Softmax layer that gives the output probabilities of all classes sum to 1. We aim to suppress the false recognized classes and assign a "Blank" label to them. Considering this, the proposed model prefers to have a "Blank" label in the output instead of a false recognized label. Results on two datasets, as shown in Table 4-7, confirmed that in a case of false recognition of the model, the recognized Softmax outputs for all classes are lower than the predefined threshold. As shown in Table 4-7, the proposed model obtains an average of recognized Softmax outputs equal to 0.98 and 0.59 on the RKS-PERSIANSIGN and ASLLVD datasets, respectively. It is worth mentioning that the average of the Softmax outputs is different from the recognition accuracy of the proposed model. The average of the Softmax outputs in each continuous sign video is the average of the maximum value of the Softmax outputs in each isolated sign. We have a higher recognition accuracy on the RKS-PERSIANSIGN dataset than the ASLLVD one due to a higher video sample instances in each class. More concretely, we have 100 and 7 sign videos for each class in the RKS-PERSIANSIGN and ASLLVD datasets, respectively. This comes from the fact that deep learning-based models generally have a better performance if they train with a large amount of data. Furthermore, there are some challenges in the similar signs, such as 'Congratulation', 'Excuse', 'Upset', 'Blame', 'Fight', 'Competition'. For example, 'Excuse' and 'Congratulation', 'Upset' and 'Blame', 'Fight' and 'Competition' signs include many similar frames. Thus, adding additional samples to these categories could help to learn more powerful features to better represent sign categories and reduce miss-classifications due to low inter-class variabilities. As future work, we aim to collect a dataset, including more realistic continuous sign videos and the corresponding isolated sign videos. Relying on this dataset, we can check the performance of the proposed model to use in a realistic scenario.